\documentclass[10pt,twocolumn,letterpaper]{article}

\usepackage{iccv}              

\usepackage{times}
\usepackage{epsfig}
\usepackage{graphicx}
\usepackage{amsmath,amssymb}
\usepackage[numbers,sort&compress]{natbib}
\usepackage{booktabs}
\usepackage[final]{microtype}
\usepackage{xcolor}
\usepackage{siunitx}
\usepackage{multirow}
\usepackage{tabularx}
\usepackage{array}
\usepackage{tikz}
\usepackage{comment}
\usepackage{mathtools}

\usepackage{enumitem}
\setlist{nosep,leftmargin=*} 

\definecolor{iccvblue}{rgb}{0.21,0.49,0.74}
\usepackage[pagebackref,breaklinks,colorlinks,allcolors=iccvblue]{hyperref}

\def\paperID{*****} 
\def\confName{ICCV}
\def\confYear{2025}

\title{Geo-NVS-w: Geometry-Aware Novel View Synthesis In-the-Wild \\ with an SDF Renderer}

\author{
Anastasios Tsalakopoulos$^{1*}$ \quad
Angelos Kanlis$^{1*}$ \quad
Evangelos Chatzis$^{1}$ \quad \\
Antonis Karakottas$^{1}$ \quad
Dimitrios Zarpalas$^{1}$ \\
$^{1}$Centre for Research and Technology Hellas (CERTH), Thessaloniki, Greece \\
{\tt\small \{tsalakop, a.kanlis, chatzise, ankarako, zarpalas\}@iti.gr} \\
{\small $^{*}$Equal contribution}
}

\begin{document}
\maketitle

\vspace{-2em}

\begin{abstract}

We introduce Geo-NVS-w, a geometry-aware framework for high-fidelity novel view synthesis from unstructured, in-the-wild image collections. While existing in-the-wild methods already excel at novel view synthesis, they often lack geometric grounding on complex surfaces, sometimes producing results that contain inconsistencies. 

Geo-NVS-w addresses this limitation by leveraging an underlying geometric representation based on a Signed Distance Function (SDF) to guide the rendering process. This is complemented by a novel Geometry-Preservation Loss which ensures that fine structural details are preserved. Our framework achieves competitive rendering performance, while demonstrating a 4–5× reduction reduction in energy consumption compared to similar methods. We demonstrate that Geo-NVS-w is a robust method for in-the-wild NVS, yielding photorealistic results with sharp, geometrically coherent details.

\end{abstract}

\addtolength{\textheight}{0.2in}

\section{Introduction}
Novel view synthesis from unconstrained, in-the-wild image collections has emerged as a cornerstone problem in computer graphics and vision. The ability to explore a 3D scene by rendering photorealistic views from any arbitrary viewpoint has profound applications in virtual reality, digital heritage, and visual effects. However, in-the-wild datasets, such as the widely used IMC-Phototourism dataset~\cite{snavely2006phototourism} consisting of tourist photos of landmarks, present a formidable challenge. They are characterized by inconsistent illumination, varying camera settings, and transient occluders such as pedestrians, vehicles, or temporary structures.

Pioneering methods like Neural Radiance Fields~\cite{mildenhall2020nerf} and its variants for in-the-wild data~\cite{martin2021nerfw,chen2022ha} have made remarkable progress. However, they inherit a fundamental limitation of volumetric density representations, which is geometric ambiguity that tends to produce semi-transparent artifacts and blur sharp architectural details.

We argue that to achieve greater photorealism and consistency in NVS, the rendering process can benefit from being explicitly guided by the scene's underlying geometry. To this end, we introduce Geo-NVS-w, a framework that uses a high-fidelity Signed Distance Function (SDF) representation~\cite{wang2021neus} as its geometric backbone. An accurate SDF provides a strong prior for surface locations, enabling our renderer to create sharp depth discontinuities and, consequently, sharper images.

\vspace{0.5em}
Our core contributions are:
\vspace{0.5em}
\begin{itemize}
  \setlength\itemsep{0.5em}     
  \setlength\topsep{0.5em}      
    \item \textbf{Octree-Accelerated feature-based volume}: We pair an SDF-guided renderer with an octree-based feature volume, interpolating features within geometry-bearing regions to preserve detail while accelerating training.
    \item \textbf{Geometry-Preservation Loss (GPL).} We introduce a novel loss that explicitly penalizes the model for incorrectly masking out geometrically significant regions as transient, ensuring textures and features remain coherent across viewpoints.
    \item \textbf{Quantified Energy Efficiency Analysis:} We instrument training with GPU power logging to quantify the energy–quality trade-off, documenting markedly lower training time and energy use.
\end{itemize}

\vspace{0.5em}
Our approach is founded on the principle that an accurate and robust geometric representation is essential for high-quality view synthesis. By grounding our rendering in an SDF, we provide a strong inductive bias for surfaces, which directly contributes to the quality of our synthesized views, especially on complex scenes. Geo-NVS-w thus demonstrates that high-quality NVS can be achieved through the synergy of advanced rendering techniques and strong geometric grounding.

\begin{figure*}[t]
    \centering
    \includegraphics[width=0.9\linewidth]{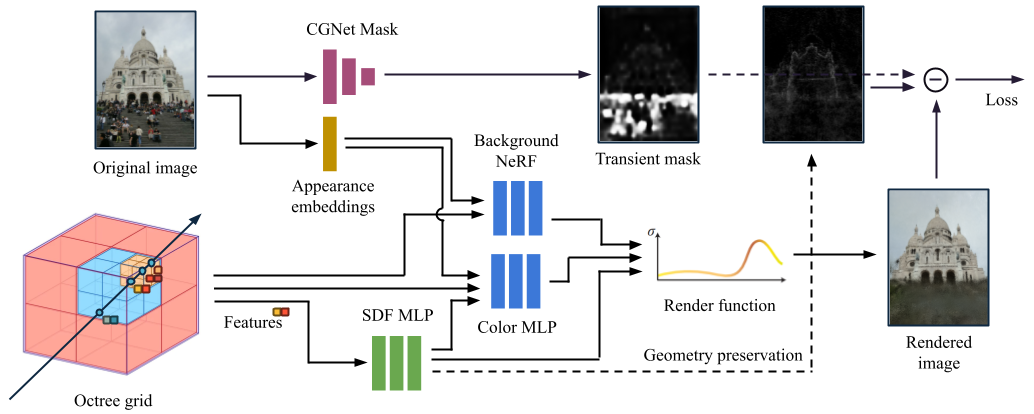}
    \caption{\textbf{Overview of the Geo-NVS-w Framework.} For a given camera ray, we march through an octree containing feature grids for the foreground (SDF-based) and background (NeRF-based). Within the foreground unit sphere, interpolated features are passed to MLPs to predict an SDF value $s(\mathbf{x})$ and a color $c(\mathbf{x}, \mathbf{d})$. We use the NeuS rendering formula to convert SDF values into alpha-compositing weights $w_i$, ensuring rendered colors are tightly coupled to the underlying surface. These weights are used to accumulate color, which is then composed with the background. Our Geometry-Preservation Loss (GPL) ensures the transient mask does not erode high-curvature details.}
    \label{fig:overview}
\end{figure*}

\section{Related Work}
Our work builds upon advances in novel view synthesis and neural surface representation.

\subsection{Novel View Synthesis}
Neural Radiance Fields (NeRF)~\cite{mildenhall2020nerf} model a scene as a continuous mapping from 5D coordinates (position and viewing direction) to volumetric density and color, enabling photorealistic rendering but requiring dense sampling and long training, and struggling with large-scale scenes. For faster rendering, PlenOctrees~\cite{yu2021plenoctrees} bake the network into an explicit octree for real-time rendering. To handle unbounded scenes, NeRF++~\cite{zhang2020nerfplusplus} uses dual MLPs to separately model foreground and background. For in-the-wild data, NeRF-W~\cite{martin2021nerfw} adds per-image appearance embeddings and a transient component for variable lighting and dynamics, while Ha-NeRF~\cite{chen2022ha} couples an appearance-hallucinating CNN with a transient 2D mask MLP.

\subsection{Neural Surface Representation}
A parallel thread targets high-quality geometry via implicit representations. NeuS~\cite{wang2021neus} introduces SDF volume rendering, ensuring the opacity-weighted color of a ray corresponds to the underlying surface. To scale, acceleration structures are crucial: Instant-NGP~\cite{mueller2022instant} employs multi-resolution hash grids for NeRFs, and Neuralangelo~\cite{2023neuralangelo} adapts them to SDFs for high-fidelity, large-scale reconstruction, highlighting the importance of numerical gradients and coarse-to-fine optimization for stabilizing SDF training on hash grids.

Geo-NVS-w unifies these lines: we adopt NeuS-style SDF rendering for geometric fidelity and grid efficiency, analogous to Neuralangelo but using an octree feature grid. Unlike reconstruction-focused work, we use this pipeline as the backbone for high-quality novel view synthesis.

\section{The Geo-NVS-w Framework}

\label{sec:methodology}
Geo-NVS-w is engineered primarily for high-fidelity novel view synthesis. Its architecture is built upon a robust geometric foundation---a Signed Distance Function (SDF)---which is key to rendering consistent views. The framework combines an efficient octree feature volume, an SDF-guided rendering process, and our novel Geometry-Preservation Loss (GPL). This geometric foundation avoids artifacts typical in NeRF renderings---such as density clouding from inconsistent ray sampling---by enforcing coherent surface geometry.

\subsection{Octree Feature Volume}
To balance performance and efficiency, we represent the scene using two separate, sparse octree feature grids: one for the foreground, modeled with an SDF, and a second for the background environment, modeled with a conventional NeRF. This dual-grid octree structure concentrates computational resources on regions containing geometry, pruning vast empty spaces. For any 3D point $\mathbf{x}$ inside the foreground's unit sphere, we query features via trilinear interpolation from the SDF grid. These features are then decoded by small MLPs to produce SDF and color values. For points outside this sphere, features are queried from the background grid and passed to a small background NeRF network.

\subsection{SDF-guided Volumetric Rendering}
Accurate surface localization is key for sharp rendering. An SDF defines the surface by the zero level set $s(\mathbf{x})=0$.

We use NeuS~\cite{wang2021neus} to link SDFs to alpha compositing. For a ray sample $\mathbf{x}_i$ with SDF $s_i=s(\mathbf{x}_i)$, the occupancy is
\begin{equation}
\alpha_i=\mathrm{sigmoid}(\zeta s_i+\beta_i)-\mathrm{sigmoid}(\zeta s_{i+1}+\beta_i),
\end{equation}
where $\zeta$ is a learned global deviation and $\beta_i=\langle\nabla s_i,\mathbf{d}\rangle \Delta t_i/2$ corrects for the angle between ray $\mathbf{d}$ and normal $\nabla s_i$. A compact \emph{Deviation Network} learns $\zeta$, and normals/derivatives use finite differences~\cite{2023neuralangelo}. Rendering weights follow discrete compositing:
\begin{equation}
w_i=T_i\alpha_i,\quad T_i=\prod_{j<i}(1-\alpha_j),
\end{equation}
and the final color $C(\mathbf{r})$ accumulates the foreground over the background NeRF. This concentrates color integration to a narrow band around the surface, producing geometrically consistent, sharp views.

\subsection{Appearance and Transient Modules}
A per-image latent appearance code is used to model variations in lighting and camera parameters, inspired by NeRF-W~\cite{martin2021nerfw}. Regarding transient occluders, we follow CR-NeRF~\cite{yang2023crossray} in using an unsupervised segmentation-based approach, where a lightweight CGNet~\cite{wu2021cgnet} is trained to generate transient object masks.

\subsection{Geometry-Preservation Loss (GPL)}
A standard challenge for in-the-wild NVS is disentangling static background from transient objects (e.g., people, cars). This is often handled with a transient mask network, which learns to down-weight pixels corresponding to dynamic elements. However, training multiple networks using a photometric loss often leads to the transient mask network removing static structure edges, leading to degradation of parts of the SDF representation, particularly those with complex geometry.

To counteract this, we introduce the \textbf{Geometry-Preservation Loss (GPL)}. The intuition is to penalize the transient mask for being active on rays that pass through geometrically significant regions. We define the loss as:
\begin{equation}
\mathcal{L}_{\text{GPL}} = \mathbb{E}_{\mathbf{r}} \left[ M(\mathbf{r}) \cdot \Phi(\mathbf{r}) \right],
\end{equation}
where $M(\mathbf{r}) \in [0,1]$ is the predicted mask value for ray $\mathbf{r}$, and $\Phi(\mathbf{r})$ is an edge indicator function derived from the foreground SDF. This indicator is designed to be high for rays intersecting geometrically complex regions. For each ray, we compute the ray-wise average of the per-sample eikonal error, $E_{\text{grad}}(r) = (\lVert \nabla s\rVert-1)^2$, and the absolute curvature estimate, $E_{\text{curv}}(r) = \lvert \Delta s\rvert$. The indicator function is then:
\begin{equation}
\Phi(\mathbf{r}) = \sigma\!\left(\lambda_{\text{grad}} E_{\text{grad}}(\mathbf{r}) + \lambda_{\text{curv}} E_{\text{curv}}(\mathbf{r})\right),
\end{equation}
where $\sigma$ is a calibrated sigmoid function. Intuitively, $\Phi(\mathbf{r})$ peaks on rays that traverse sharp edges or areas of high curvature. By multiplying the mask value with this indicator, our geometry preservation loss encourages the mask to remain zero (i.e., fully foreground) for rays that are crucial for defining the scene's structure. This is designed to preserve sharp features, leading to visibly sharper and more coherent novel views. We set the loss weights $\lambda_{\text{GPL}} = 0.05$, $\lambda_{\text{grad}} = 10.0$, and $\lambda_{\text{curv}}=2.0$ in all our experiments.

\subsection{Learning Objective}
The complete framework is trained end-to-end by minimizing a composite loss:
\begin{equation}
\begin{split}
\mathcal{L} ={} & \lambda_{\text{rgb}}\mathcal{L}_{\text{rgb}} + \lambda_{\text{eik}}\mathcal{L}_{\text{eik}} + \lambda_{\text{curv}}\mathcal{L}_{\text{curv}} \\
                & + \lambda_{\text{mask}} L_{\text{reg}} + \lambda_{\text{GPL}}\mathcal{L}_{\text{GPL}} + \lambda_{\text{lps}}\mathcal{L}_{\text{lps}}.
\end{split}
\end{equation}
The key components include: the primary photometric loss $\mathcal{L}_{\text{rgb}}$ (L1 difference); an eikonal loss $\mathcal{L}_{\text{eik}} = (\lVert \nabla s\rVert-1)^2$; our novel $\mathcal{L}_{\text{GPL}}$; the Lipschitz loss $\mathcal{L}_{\text{lps}}$~\cite{liu2022learning}, on the color network, following prior work~\cite{rosu2022permuto}; a curvature loss $\mathcal{L}_{\text{curv}} = \lvert \Delta s\rvert$; and an off-surface penalty. The regularized transient mask, $L_{\text{reg}}$, combines the transient CNN mask with penalties for intensity and bimodality to encourage a clean separation of static and dynamic elements. During a warm-up phase, we also use a sphere-initialization loss, $\mathcal{L}_{\text{sphere}}$, to provide a stable initial geometry.

\begin{figure}[t]
  \centering
  \setlength{\tabcolsep}{1pt} 
  \renewcommand{\arraystretch}{1} 
  \begin{tabular}{cccc}
    \includegraphics[width=0.18\linewidth]{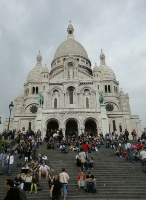} &
    \includegraphics[width=0.18\linewidth]{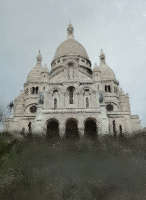} &
    \includegraphics[width=0.18\linewidth]{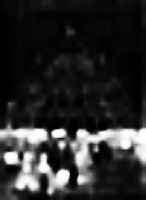} &
    \includegraphics[width=0.18\linewidth]{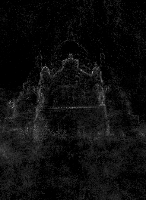} \\
    \includegraphics[width=0.18\linewidth]{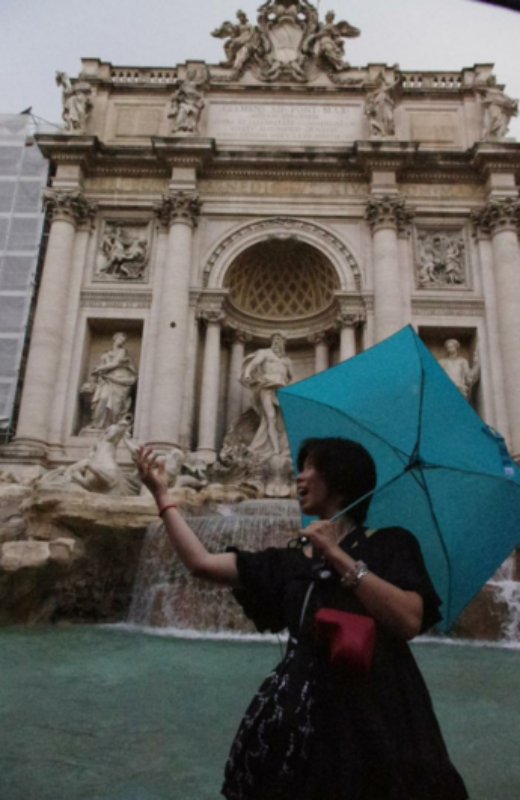} &
    \includegraphics[width=0.18\linewidth]{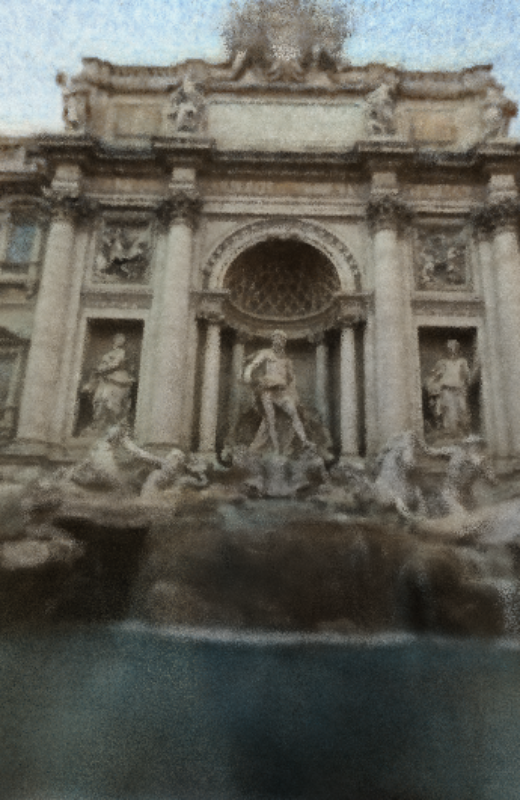} &
    \includegraphics[width=0.18\linewidth]{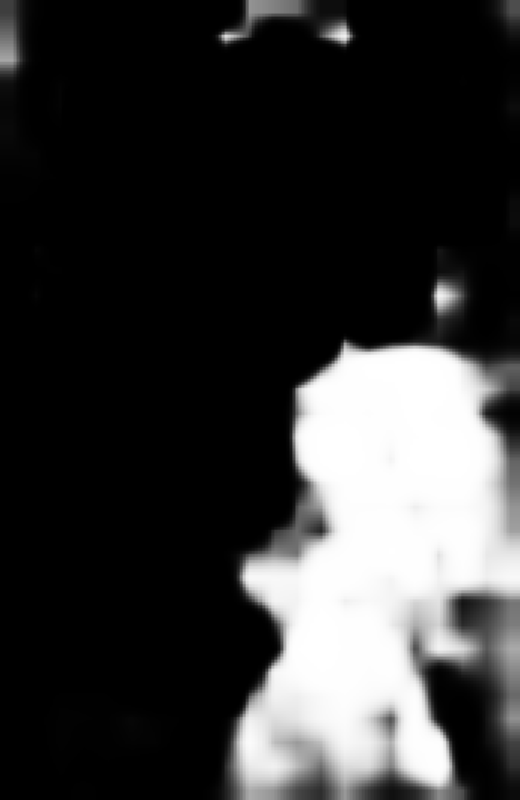} &
    \includegraphics[width=0.18\linewidth]{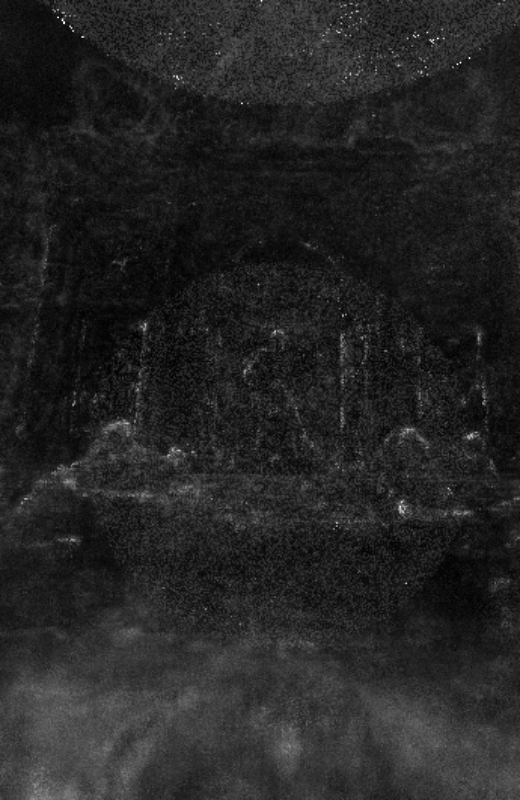} \\
    \includegraphics[width=0.24\linewidth]{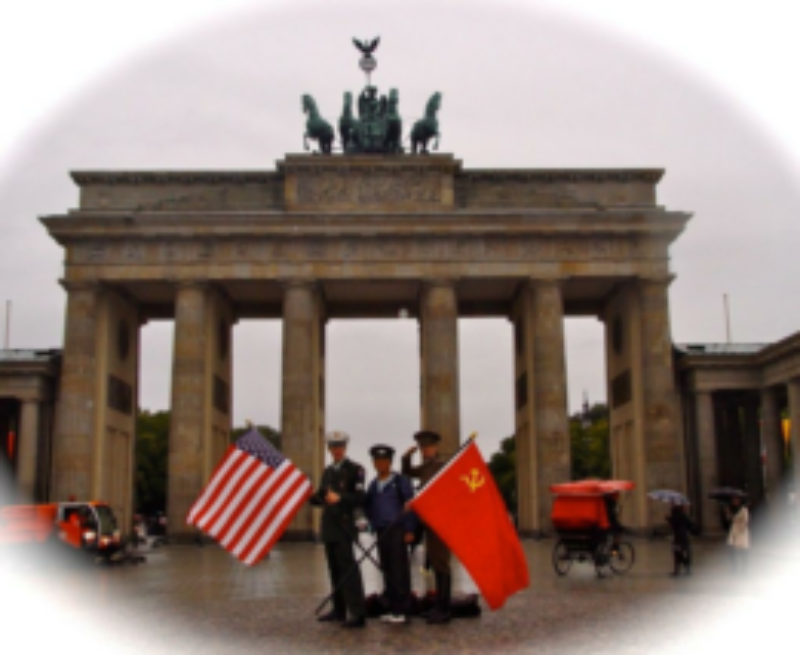} &
    \includegraphics[width=0.24\linewidth]{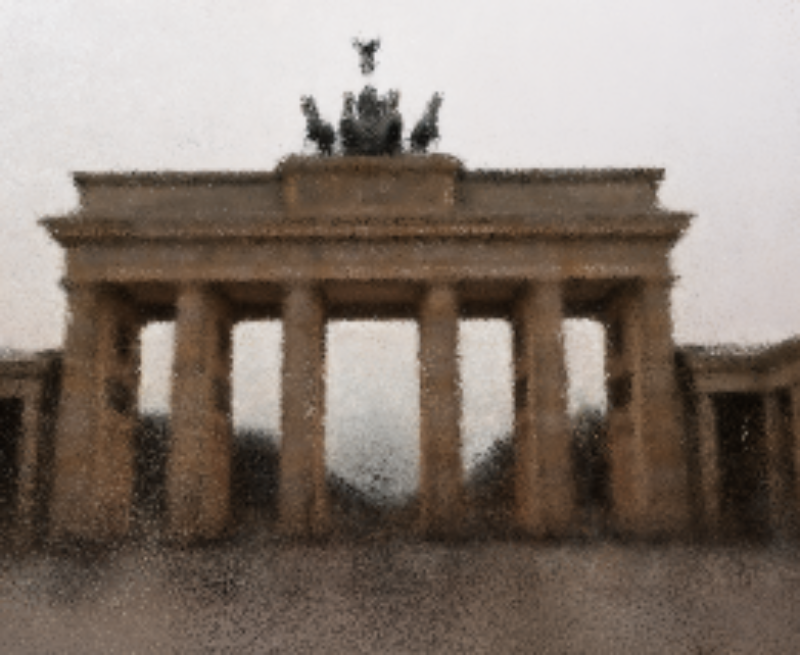} &
    \includegraphics[width=0.24\linewidth]{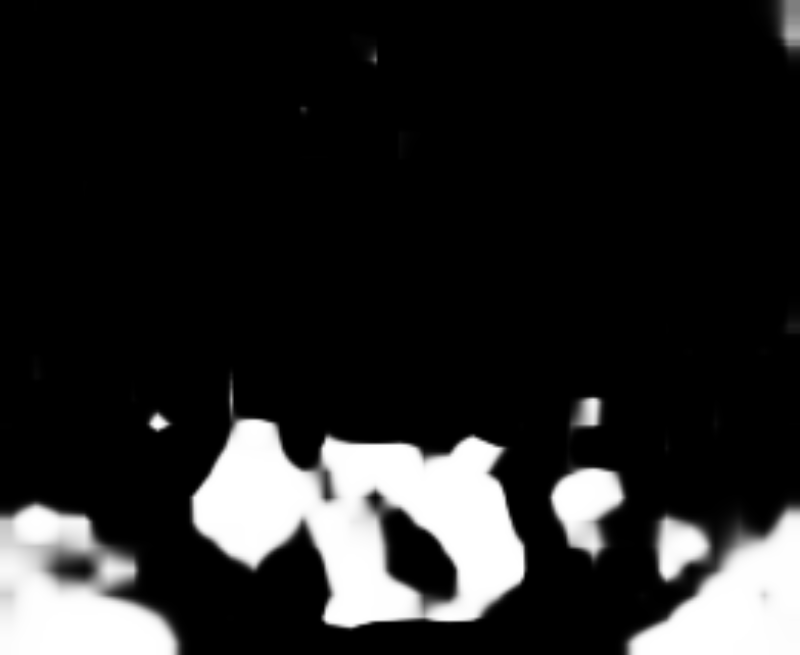} &
    \includegraphics[width=0.24\linewidth]{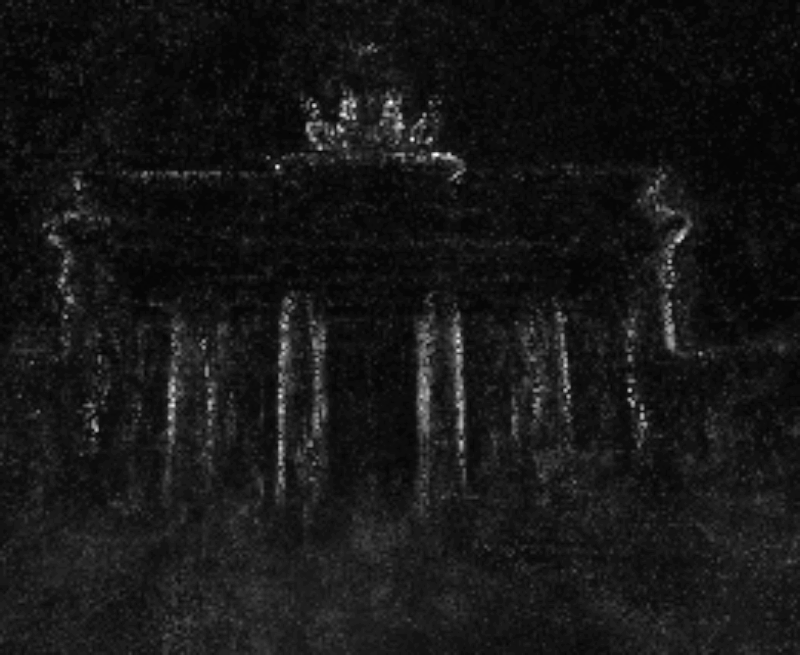} \\
    (a) & (b) & (c) & (d) \\
  \end{tabular}
  \caption{\textbf{Qualitative results on Phototourism scenes.} Top to bottom: Sacré-Cœur, Trevi Fountain, Brandenburg Gate. (a) Input image. (b) Rendered result with Geo-NVS-w from the same viewpoint and appearance embedding. (c) Estimated transiency mask. (d) Visualization of the geometry-preservation map (accumulated SDF gradients along each ray).}
  \label{fig:teaser}
\end{figure}
\section{Experiments}
We evaluate Geo-NVS-w on all four currently available scenes from the IMC-PT dataset\cite{snavely2006phototourism} also used in NeRF-W and Ha-NeRF,  as the rest have been removed due to data inconsistencies \cite{yi2020imc_dataset}. We also view CR-NeRF ~\cite{yang2023crossray} as complementary and plan to include a controlled comparison in future work.

\subsection{Implementation Details}
Our model is implemented in PyTorch~\cite{paszke2019pytorch} and trained on a single NVIDIA A10G GPU using mixed-precision computation to accelerate training. We measure energy consumption by incorporating GPU power measurement directly into our training pipeline, logging cumulative usage over time.

\begin{table}[t]
  \centering
  \caption{\textbf{Novel View Synthesis (NVS) quality.} We report PSNR ($\uparrow$), SSIM ($\uparrow$), and LPIPS ($\downarrow$). Geo-NVS-w achieves strong results on all IMC-PT dataset scenes.}
  \label{tab:nvs_results}
  \setlength{\tabcolsep}{3pt}
  \renewcommand{\arraystretch}{0.8}
  \small

  \begin{tabular*}{0.8\linewidth}{@{\extracolsep{\fill}} lccc @{}}
    \toprule
    Method & PSNR $\uparrow$ & SSIM $\uparrow$ & LPIPS $\downarrow$ \\
    \midrule
    \multicolumn{4}{c}{\textbf{Brandenburg Gate}} \\
    Ha-NeRF & 23.45 & 0.811 & 0.247 \\
    NeRF-W & 23.98 & 0.915 & 0.198 \\
    \textbf{Ours} & \textbf{25.40} & \textbf{0.944} & \textbf{0.158} \\
    \addlinespace[0.4em]
    \multicolumn{4}{c}{\textbf{Sacr\'e-C\oe ur}} \\
    Ha-NeRF & \textbf{25.40} & \textbf{0.877} & \textbf{0.124} \\
    NeRF-W & 25.11 & 0.859 & 0.141 \\
    \textbf{Ours} & 23.23 & 0.850 & 0.160 \\
    \addlinespace[0.4em]
    \multicolumn{4}{c}{\textbf{Trevi Fountain}} \\
    Ha-NeRF & 22.15 & 0.695 & \textbf{0.117} \\
    NeRF-W & 23.01 & 0.751 & 0.109 \\
    \textbf{Ours} & \textbf{24.50} & \textbf{0.831} & 0.203 \\
    \addlinespace[0.4em]
    \multicolumn{4}{c}{\textbf{Taj Mahal}} \\
    Ha-NeRF & 22.72 & 0.767 & 0.301 \\
    NeRF-W & \textbf{25.15} & 0.833 & 0.195 \\
    \textbf{Ours} & 24.19 & \textbf{0.860} & \textbf{0.194} \\
    \midrule
    \multicolumn{4}{c}{\textbf{Average}} \\
    Ha-NeRF & 23.43 & 0.788 & 0.197 \\
    NeRF-W & 24.31 & 0.840 & \textbf{0.161} \\
    \textbf{Ours} & \textbf{24.33} & \textbf{0.871} & 0.179 \\
    \bottomrule
  \end{tabular*}
\end{table}

\subsection{Novel View Synthesis Benchmarks}

Our method matches or exceeds baselines \cite{chen2022ha,martin2021nerfw} on several metrics, as shown in \cref{tab:nvs_results}, while achieving superior image quality metrics in most scenes. \cref{fig:energy_psnr_tradeoff} corroborates these findings, illustrating that our method delivers enhanced visual fidelity at higher processing speed and reduced energy expenditure.

\subsection{Energy Analysis}
\begin{figure}[t]
    \centering
    \includegraphics[width=\linewidth]{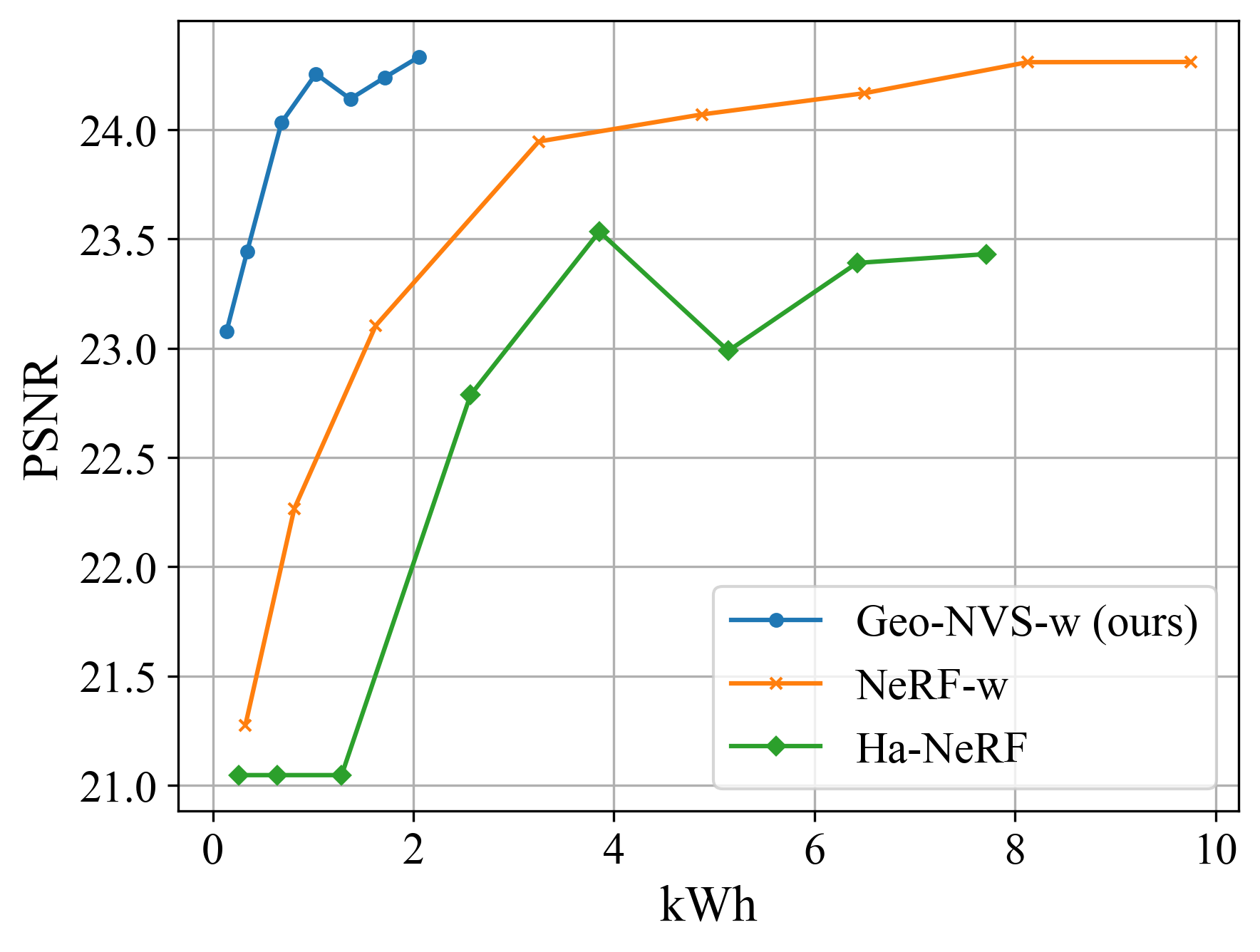}
    \caption{\textbf{Energy vs.\ quality trade-off.} Geo-NVS-w achieves high PSNR with significantly less training time and cumulative energy consumption (kWh) compared to baseline NeRF-W, making it a more efficient and scalable solution.}
    \label{fig:energy_psnr_tradeoff}
\end{figure}

Beyond raw performance, practical usability and scalability depend on computational efficiency. As shown in \cref{fig:energy_psnr_tradeoff}, Geo-NVS-w is not only faster but also more energy-efficient. Our method completes a 300{,}000-iteration run using approximately \SI{2.05}{kWh}, whereas NeRF-W consumes \SI{9.7}{kWh} and Ha-NeRF consumes \SI{7.71}{kWh}. Beyond achieving superior geometric fidelity, our SDF-based approach also demonstrates efficiency gains, reaching peak PSNR on our runs using approximately \SI{2.05}{kWh}, a beneficial effect of an architecture that allows for significant downsizing of the MLPs.

\section{Conclusion}
We presented Geo-NVS-w, a framework advancing in-the-wild novel view synthesis by placing geometric preservation at the core of the rendering process. Incorporating a Signed Distance Function into our architecture and introducing a Geometry-Preservation Loss, our method produces renderings with consistent sharpness and geometrical coherence. The framework offers a strong, geometry-consistent alternative with favorable efficiency–quality trade-offs compared to prior approaches. While newer paradigms such as Gaussian Splatting have emerged, our findings confirm that geometry-aware methods remain effective for high-fidelity, in-the-wild view synthesis.

Future work could explore integrating 3D Gaussian Splatting techniques~\cite{kerbl20233d} within in-the-wild settings~\cite{kulhanek2024wildgaussians,zhang2024gaussian}, leveraging their speed advantages to further reduce rendering latency and computational overhead while maintaining the surface coherence enabled by our SDF methodology, as presented in~\cite{li2025gs-sdf}.

\section{Acknowledments}
This research has been supported by the European Commission funded program XReco, under Horizon Europe Grant Agreement 101070250. The computational resources were granted with the support of GRNET.

{\small
\bibliographystyle{ieeenat_fullname}
\bibliography{main}
}

\end{document}